\documentclass[english]{sobraep}
\usepackage[misc]{ifsym}
\usepackage{amsmath}
\usepackage{amsfonts}
\usepackage[justification=centering]{caption}
\usepackage{hyperref}
\usepackage{fontawesome}

\title{RealNet: Combining Optimized Object Detection with Information Fusion Depth Estimation Co-Design Method on IoT }

\author{Zhuohao Li \textsuperscript{\Letter}\thanks{edith_lzh@sjtu.edu.cn}, Fandi Gou$^{2}$, Qixin De$^{3}$, Leqi Ding$^{4}$, Yuanhang Zhang$^{5}$, Yunze Cai$^{6}$\\
	\normalsize $^{1 2 3 4 5 6}$Shanghai Jiao Tong University, Shanghai, China\\
	\normalsize $^{1}$ edith\_lzh@sjtu.edu.cn, $^{2}$fandi@sjtu.edu.cn, $^{3}$qxde@sjtu.edu.cn, \\$^{4}$leqi@sjtu.edu.cn, $^{5}$yhzhang@sjtu.edu.cn, $^{6}$yzcai@sjtu.edu.cn
}

\begin{document}

\maketitle

\begin{abstract}
	Depth Estimation and Object Detection Recognition play an important role in autonomous driving technology under the guidance of deep learning artificial intelligence~\cite{heisley1991autodriving}. We propose a hybrid structure called RealNet: a co-design method combining the model-streamlined recognition algorithm, the depth estimation algorithm with information fusion, and deploying them on the Jetson-Nano for unmanned vehicles with monocular vision sensors. We use ROS for experiment. The method proposed in this paper is suitable for mobile platforms with high real-time request. Innovation of our method is using information fusion to compensate the problem of insufficient frame rate of output image, and improve the robustness of target detection and depth estimation under monocular vision.Object Detection is based on \href{https://github.com/ultralytics/yolov5}{YOLO-v5}( \href{https://github.com/ultralytics/yolov5}{https://github.com/ultralytics/yolov5}) ~\cite{glenn_jocher_2022_6222936}. We have simplified the network structure of its DarkNet53~\cite{biddle2002darknet} and realized a prediction speed up to 0.01s. Depth Estimation is based on the VNL Depth Estimation, which considers multiple geometric constraints in 3D global space.~\cite{yin2019enforcing} It calculates the loss function by calculating the deviation of the virtual normal vector VN and the label, which can obtain deeper depth information. We use PnP fusion algorithm to solve the problem of insufficient frame rate of depth map output. It solves the motion estimation depth from three-dimensional target to two-dimensional point based on corner feature matching, which is faster than VNL calculation. We interpolate VNL output and PnP~\cite{lepetit2009epnp} output to achieve information fusion. Experiments show that this can effectively eliminate the jitter of depth information and improve robustness. At the control end, this method combines the results of target detection and depth estimation to calculate the target position, and uses a pure tracking control algorithm to track it.
\end{abstract}

\begin{keywords}
	Virtual Normal Loss, Perspective-n-Point, YOLO-v5, Kalman Filter, Pure Pursuit
\end{keywords}

\let\thefootnote\relax\footnotetext{\hspace*{-5mm}\textsuperscript{\Letter} corresponding author\\
Open source project: \faGithub~\href{https://github.com/edithlzh/VNL_Estimation}{https://github.com/edithlzh/VNL\_Estimation}}



\section{INTRODUCTION}

With the rapid development of artificial intelligence technology represented by deep learning in recent years, computer vision has become an important research method in the field of image processing and pattern recognition.~\cite{forsyth2011computer}~\cite{jarvis1983perspective}~\cite{forsyth2011computer} Object Detection and Depth Estimation are both classic problems in the field of computer vision. They are widely used in robotics (Robotics Science), augmented reality (AR), ~\cite{tzafestas2018synergy}three-dimensional reconstruction (3D Rebuilding), automatic driving (Auto Driving) and other fields. With the continuous iterative improvement of hardware computing power,~\cite{robo} these new technologies are gradually reducing costs and entering our lives, so as to promote the progress of human science and technology and the improvement of people's quality of life.

\subsection{Related Work}
\textbf{Object Detection}The main task of target detection is to determine whether a certain area of the image contains the object to be recognized according to the input image. Recognition is the ability of the program to recognize the object.~\cite{obj1} Recognition usually only processes the area of the detected object.~\cite{obj2} In computer vision, there are many target detection and recognition technologies,~\cite{obj3} such as gradient histogram (Histogram of Oriented Gradient, HOG)~\cite{Histograms}, image Pyramid~\cite{adelson1984pyramid}, sliding window (Sliding Window)~\cite{lee2001sliding}. All these target detection technologies realize detection and recognition with specific detection boxes and confidence scores of recognition information. ImageNet~\cite{deng2009imagenet} Large-scale Visual Recognition Challenge (ILSVRC) is an annual event for target detection. In the 12-year history, numerous classic target detection algorithms and neural network models have emerged,~\cite{krizhevsky2012imagenet} affecting almost all modern artificial intelligence problems not only target detection and recognition.

\textbf{Depth Estimation}The main task of depth estimation is to realize the prediction from two-dimensional image to depth image and restore the depth dimension information lost by ordinary cameras in the imaging process.~\cite{torralba2002depth} At present, there are many devices in the market that can directly obtain depth, usually by emitting ultrasonic waves or laser and other physical rays to complete distance measurement, but most of them are expensive, and most of the time depth information is not our most concern. ~\cite{godard2017unsupervised}At this level, the depth estimation in computer vision can achieve good depth information restoration by means of simple algorithms and only increasing the calculation force. It has a wide range of application prospects in scenes with high real-time requirements such as mobile devices in the future. Depth estimation algorithms are usually divided into monocular camera-based estimation and binocular camera-based estimation. The latter can use physical parallax to perform pixel point correspondence calculations through geometric method stereo matching. The accuracy is not as good as that of devices that can directly obtain depth. At the same time, the overhead is larger than monocular cameras and is more restrictive. Especially for low-texture scenes, the matching effect is not ideal, so it is not widely used. Compared with panoramic vision sensor and binocular vision sensor, monocular vision sensor has the advantages of simple structure, flexible movement and easy calibration. Therefore, the depth estimation algorithm based on monocular camera has a broad prospect, which can greatly reduce the use cost of the sensor.

\subsection{Conclusion}
Conclusion In this work, we propose an innovative method to apply the target detection algorithm based on YOLO-v5 and the depth estimation algorithm based on VNL to the mobile platform, and optimize the algorithm according to the real-time required by the work application scene. On this basis, the algorithm is deployed to the Jetson-Nano platform and the ROS system is used to control the trolley. We have designed experiments to verify the performance of this method. On the surface of the experimental results, our method has increased the prediction speed by three times compared with the unoptimized model in a specific experimental scenario, the accuracy has been improved by 13\%, and the anti-interference ability is excellent. It has good performance and robustness.

\section{Object Detection Optimization}
At present, the target detection algorithm is mainly divided into two methods: one is a two-stage method based on Faster R-CNN,~\cite{girshick2015fast} which is mainly divided into two parts, one is to generate candidate boxes through a special module, and the other is to find the prospect and adjust the bounding box. The other is the One-Stage algorithm represented by SSD~\cite{liu2016ssd} and YOLO, which is directly based on anchor to classify and adjust the bounding box. The two methods have their own advantages. The Two-Stage method has higher accuracy, but the detection speed is relatively slow. The speed of One-Stage method is obviously better than that of Two-Stage method.~\cite{tian2019fcos} According to the needs, due to the high real-time requirements of this project, we chose YOLOv5 algorithm of YOLO series for target detection.

YOLO series is widely used in unmanned driving and other target detection scenarios due to its advantages of high detection accuracy and fast reasoning speed. Different from the traditional One-Stage method, YOLO does not need the region proposal stage, but directly generates the category probability and position coordinate value of the object. Therefore, YOLO can directly obtain the final detection result after one stage, so it has a faster detection speed~\cite{laroca2018robust}. YOLO model defines the target detection problem as a regression fitting problem (Regression Problem) of image classification and object positioning frame. The algorithm divides the image into S×S grids, and the neural network predicts each grid separately to obtain a (B × 5C) dimensional tensor composed of B boxes (including the probability P of objects in the grid, bounding box size height H, width W and center position (X, Y) and C category confidence (single thermal coding, also one-hot encoding). After obtaining the high-dimensional feature coding of the image (as shown in fig. 1, which is a feature map of (7 × 7 × 1024), the feature is directly mapped to the prediction result tensor S × S × (B × 5C) through a nonlinear fully connected layer.

\subsection{YOLO Series}
At present, YOLO has developed to the 5th generation, and its main improved modules are:
\subsubsection{Anchor mechanism:}the detection box changes from the previously predicted absolute position information (X, Y, H, W) to the deviation ratio (0-1) of the prediction relative to the priori candidate box. Due to the characteristics of neural network normalization, the values in the range of 0-1 will be easier to handle, so the accuracy of the detection box is improved.
\subsubsection{Feature encoder}The backbone (Backbone) of the network used by YOLO is changed from GoogleNet to DarkNet53 Focus CSP network, which can better extract image features and improve the effect of classification and positioning.
\subsubsection{Detection head}The features of the Backbone code are down-sampled twice, and then large and small objects are detected according to the obtained feature maps of three different receptive fields, thus effectively improving YOLO's detection capability for small targets.
\subsubsection{Loss function}upgrading from the original IOU loss function to GIOU loss function~\cite{xu20193d}~\cite{rezatofighi2019generalized}, the characterization ability of bounding box prediction deviation has been greatly improved.
\subsubsection{Data enhancement}The Mosaic data enhancement algorithm~\cite{irani1995mosaic} is proposed, which increases the amount of trainable data, improves the generalization ability of the model, The iteration upgrade of each generation on the Backbone and detection head is shown in the following table.

\begin{figure}[H]
    \centering
    \includegraphics[width=8cm]{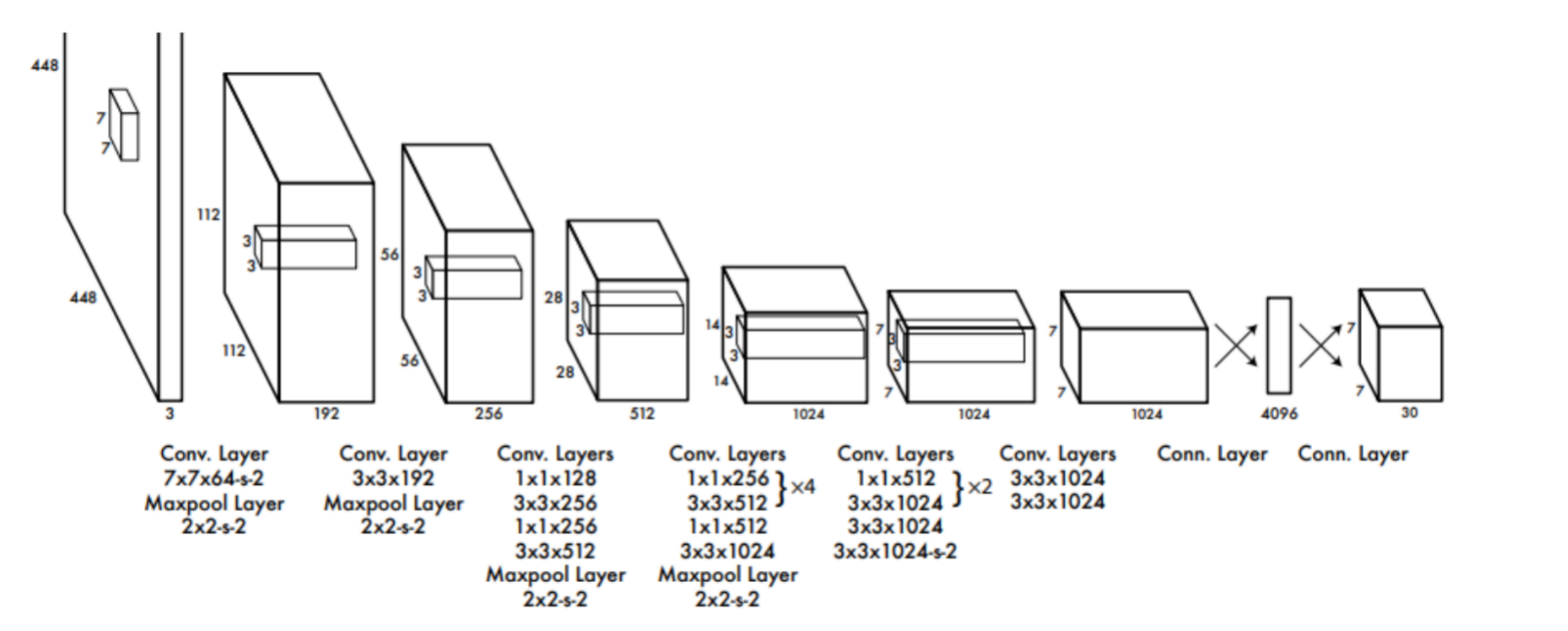}
    \centering {\caption{Network Structure of YOLO}}
    \centering{\label{fig:my_label}}
\end{figure}

\begin{table}[H]
	\centering
	\caption{\textbf{YOLO Backbones}}
	\footnotesize
	\setlength{\tabcolsep}{0pt}
	\begin{tabular}{ccc}
		\toprule [1.3pt]	
		\multicolumn{3}{c}{ \textbf{From YOLO-v1 to YOLO-v5} } \\
		\hline
		\multirow{2}{1.1cm}{\centering \textbf{YOLO Versions}} & \multirow{2}{*}{\textbf{Backbones}} &
		\multirow{2}{*}{\textbf{Head}}\\
		&  &  \\		
		\hline
		YOLO-v1&GoogleNet&$YOLO: Fc\rightarrow7\*7\*(5+5+20)$\\
		\hline
		\multirow{2}{*}{YOLO-v2} &
		\multirow{2}{2cm}{\centering DarkNet19}  &
		\multirow{2}{5cm}{\centering $Passthrough;$\\
                            $Conv \rightarrow 13*13*5*(5+20)$}\\
        & &   \\

		\hline
		\multirow{4}{*}{YOLO-v3/4/5} &
		\multirow{4}{2cm}{\centering DarkNet53}  &
		\multirow{4}{5cm}{\centering $Conv\rightarrow13*13*5*(5+80)$\\ $Conv\rightarrow26*26*5*(5+80)$\\ $Conv\rightarrow52*52*5*(5+80)$} \\
		& &   \\
		& &   \\
		& &   \\

		\hline
		\bottomrule[1.3pt]
	\end{tabular} \label{table:TableI}
\end{table}

The loss function trained by YOLO uses the MSE method, where the first term is the error term of the center coordinate of the bounding box, the second term is the error term of the height and width of the bounding box, the third term is the confidence of the bounding box containing the target The confidence error term, the fourth term is the confidence error term of the bounding box that does not contain the target, and the last term is the classification error term of.

$$
\begin{aligned}
    Loss &= \lambda_{coord}\sum^{S^2}_{i=0}\sum^{B}_{j=0}I_{ij}^{obj}[(x_i-\hat{x_i})^2+(y_i-\hat{y_i})^2] \\
    &+\lambda_{coord}\sum^{S^2}_{i=0}\sum^{B}_{j=0}I_{ij}^{obj}[(\sqrt{w_i}-\sqrt{\hat{w_i}})^2+(\sqrt{h_i}-\sqrt{\hat{h_i}})^2]\\
    &+\sum^{S^2}_{i=0}\sum^{B}_{j=0}I_{ij}^{obj}(C_i-\hat{C_i})^2+\lambda_{noobj}\sum^{S^2}_{i=0}\sum^{B}_{j=0}I_{ij}^{noobj}(C_i-\hat{C_i})^2 \\
    &+\sum^{S^2}_{i=0}I^{obj}_i\sum_{c \in classes}(p_i(c)-\hat{p_i(c)})^2
\end{aligned}
$$

However, this method has some defects, that is, the square loss cannot well measure the overlapping area of the prediction box and the real box. In order to solve this problem, YOLO gradually upgraded the loss function of the prediction box to IOU and GlOU in subsequent versions. The latter is the loss function used in YOLOv5, which selects the smallest box that frames both the real box and the prediction box, subtracts the area of the prediction box and the real box from the area of the box, and then compares the area of the selected box to reflect the distance between the real box and the prediction box. The loss calculation formula is as follows:

$$
\begin{aligned}
    & x_1^c = min(x_1^B, x_1^{Bgt})\\
    & x_2^c = min(x_2^B, x_2^{Bgt})\\
    & y_1^c = min(y_1^B, y_1^{Bgt})\\
    & y_2^c = min(y_2^B, y_2^{Bgt})\\
\end{aligned}
$$

$$
\begin{small}
\begin{aligned}
    & GloU(B,B_{gt})  = IoU(B,B_{gt}) - \cfrac{\mid C-(B\cup B_{gt}) \mid }{\mid C \mid} \\
   &L_{GIoU(B,B_{gt})}=1-GIoU(B,B_{gt})  =1-IoU(B,B_{gt})-\cfrac{\mid C-(B\cup B_{gt}) \mid }{\mid C \mid}
\end{aligned}
\end{small}
$$

In the experiment, we simplified the YOLO-v5 source code to obtain a model reasoning package that can run directly. When collecting data, we control the moving target to move, shoot the video of the target movement through the camera of the car, and obtain pictures from the video. When training the model, we used a variety of data enhancement~\cite{irani1995mosaic} methods before labeling the material, such as geometric transformation~\cite{gielis2003generic}, grayscale transformation, mosaic data enhancement, and for continuous timeline pictures taken, randomly disrupting the sequence during training, etc., to improve the reliability of the final model at the training set level. We use Mosaic data enhancement method when training the model, and draw lessons from the idea in YOLOX~\cite{ge2021yolox}, turn off Mosaic enhancement in 15 epoch before the end of training, so that the detector can avoid the influence of inaccurate labeling boxes and complete the final convergence under the data distribution of natural pictures.

In addition, we chose yolov5\_s.pt with the fastest detection speed as our pre-training model. Since the Jetson-Nano is a single chip microcomputer based on Arm architecture, the anaconda python environment cannot be used, and the ROS system does not support python3, which brings challenges to the experiment. We installed python environment in virtual environment, configured pytorch framework and CUDA GPU accelerator, and finally successfully deployed YOLOv5 algorithm. The training diagram and the reasoning results on the Jetson-Nano are shown in Figure 2.

\begin{figure}[htbp]
\centering
\subfigure[before]{
\begin{minipage}[t]{0.5\linewidth}
\centering
\includegraphics[width=1.5in]{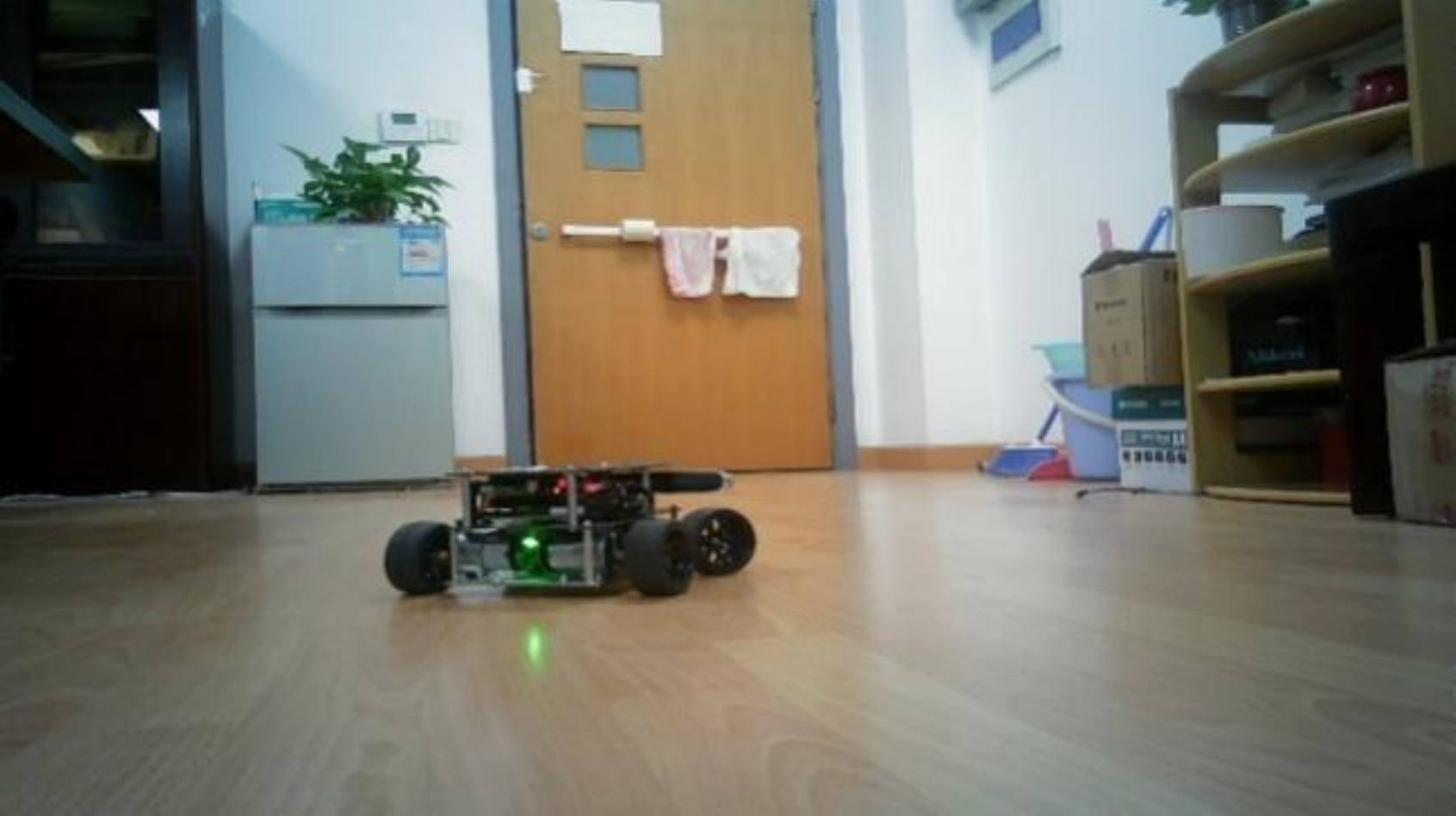}
\end{minipage}%
}%
\subfigure[after]{
\begin{minipage}[t]{0.5\linewidth}
\centering
\includegraphics[width=1.5in]{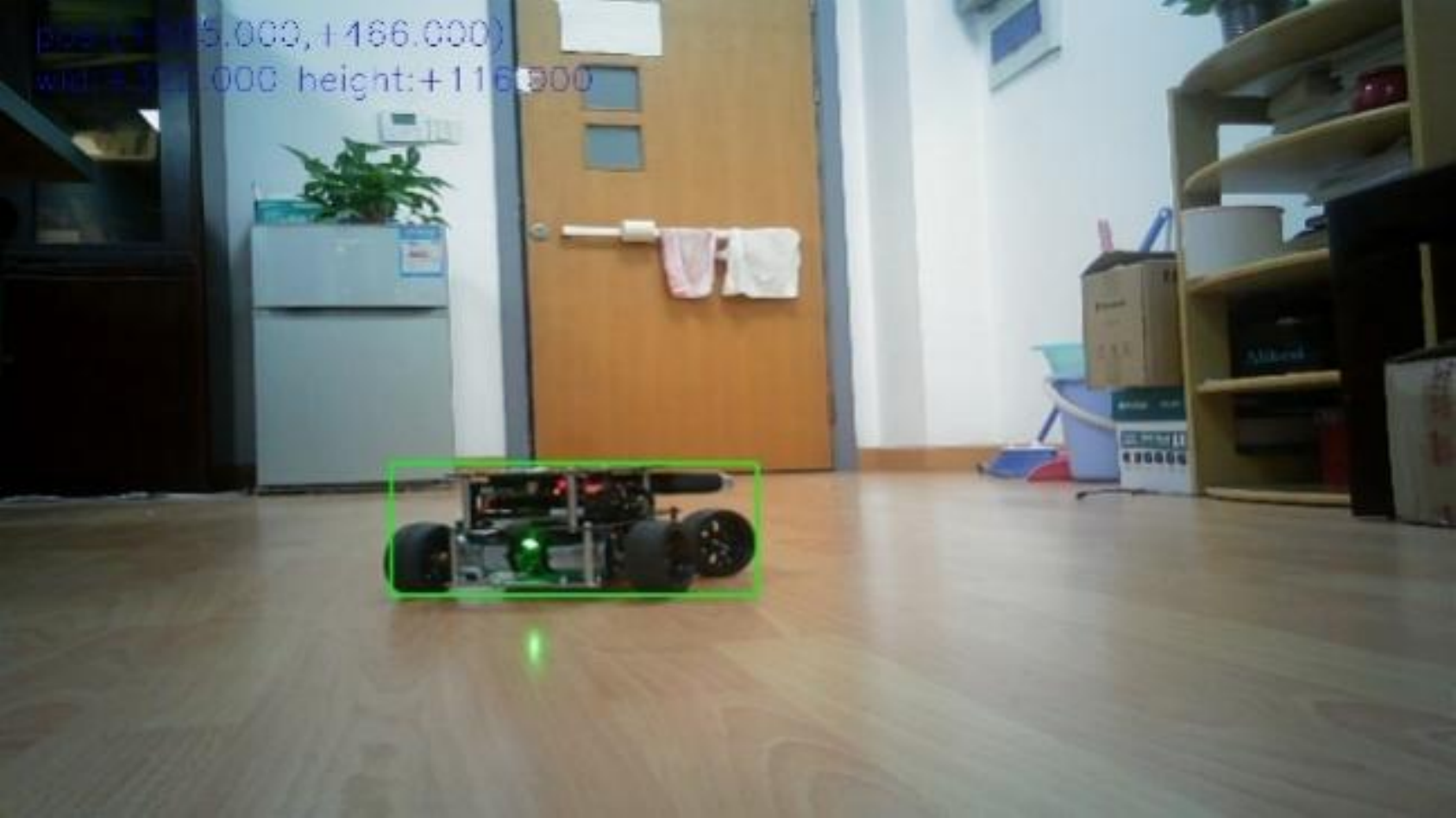}
\end{minipage}%
}%
\centering
\caption{Pictures before and after reasoning}
\end{figure}

\section{Depth Estimation}

\subsection{VNL Depth Estimation}

Traditional monocular camera depth estimation is mainly based on point-to-point (point-wise) method. The two-dimensional plane image pixels captured by the camera are used as input, and a deep convolutional neural network (DCNN)~\cite{li2018deep}~\cite{laina2016deeper}~\cite{guo2018learning}~\cite{fu2018deep}~\cite{eigen2014depth}~\cite{eigen2015predicting} is used to realize the mapping of two-dimensional information to three-bit information. Specifically, it is divided into "positive methods" and "negative methods". The former uses some auxiliary optical information to help predict, such as the "coded patterns" proposed by W.Yin and others in CVPR2017~\cite{yin2017high} Actively seek additional information assistance. The latter focuses completely on the image itself, such as the method proposed by Y.Cao et al. in TCSVT2017~\cite{cao2017estimating}, D.Eigen et al. in CVPR2015~\cite{eigen2014depth} and NeuIPS2014~\cite{eigen2015predicting}. The latter is completely an image recognition problem, but because two-bit images can directly mine less three-dimensional information, the performance has not been particularly ideal. Thanks to the proposal of DCNN such as ResNet~\cite{he2016deep}, the new network allows more massive data processing, and some large-scale feature methods and attention mechanisms have also been introduced. Liu et al. in PAMI2016~\cite{liu2015learning} proposed CRF, a method of co-processing information around a single pixel, which is considered a breakthrough, making depth estimation technology progress from point-to-point to pair-wise. Along this line of thinking, Chen et al. in CRR2018~\cite{chen2018rethinking} proposed a method to combat the generated network (GAN)~\cite{goodfellow2014generative} so that the network has context-aware and patch-wise processing. However, all this work only focuses on the local depth information itself, not from a global perspective, that is, the entire 3D space. However, due to the high noise of consumer cameras, the fluctuation of adjacent information cannot be predicted, which will have a great impact on performance.

An article published in ICCV2019 by Wei Yin et al.~\cite{yin2019enforcing} proposed a high-order geometric feature constraint (High-order geometric constraints) for depth estimation, which improves the accuracy and robustness of depth estimation. This is a supervised monocular depth estimation method that simplifies the sub-model construction process required for previous depth estimation. This method mainly proposes a geometric constraint considered in 3D global space, called Virtual Normal Loss(VNL), which is a deeper depth information, rather than simply coming directly from the depth parameter. The VNL method first constructs a depth map from the traditional point-wise method, reconstructs a 3D cloud point map on the basis of this estimated depth map, and then randomly selects three distant and linearly independent points from the cloud point map to construct a virtual plane. The normal vector of the virtual plane is the virtual normal vector VN. The loss function is determined by the difference between the VN and the true normal vector. This method can obtain not only depth information, but also other information such as location information, including coordinates. The framework of this method is shown in Figure 3, and each part is described in detail below.

\begin{figure}[H]
    \centering
    \includegraphics[width=8cm]{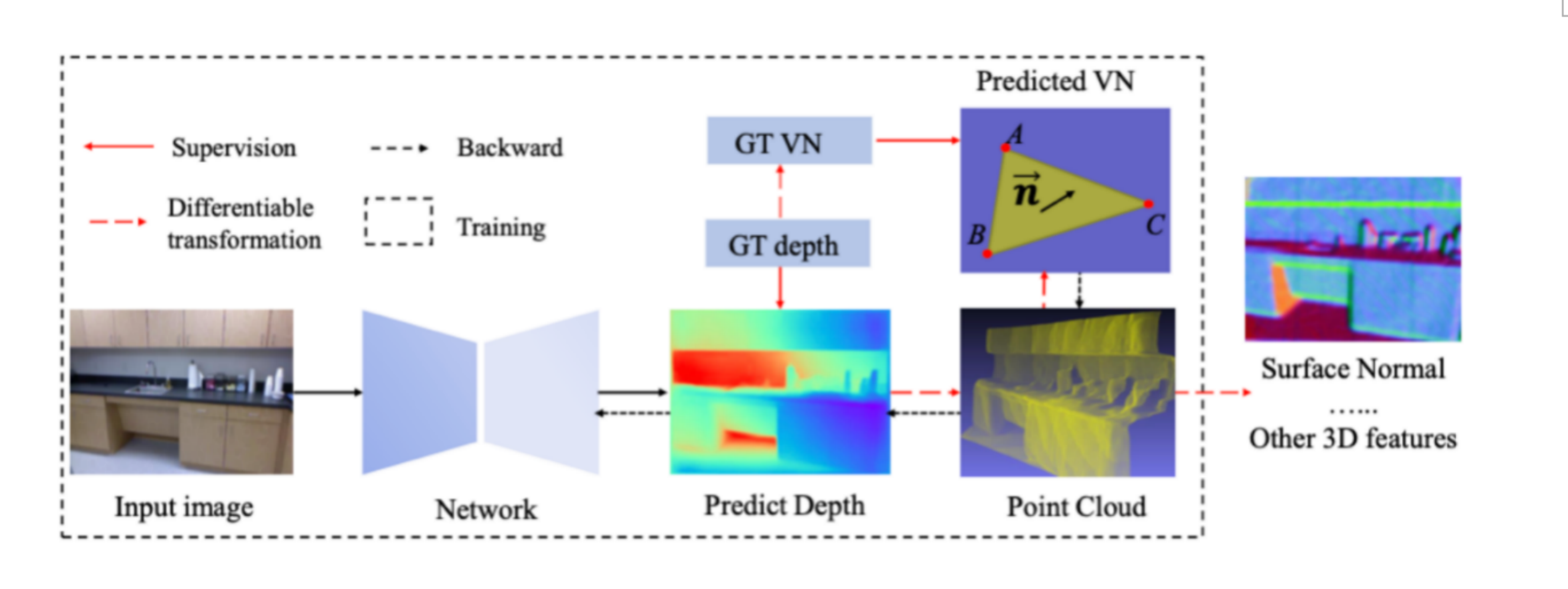}
    \caption{VNL Flow}
    \centering{\label{fig:my_label}}
\end{figure}

First, the input image is generated through the point-wise network to generate an "estimated depth map", then a virtual normal vector of the plane is generated based on this estimated map, and the cloud point map is reconstructed under the supervision of VN. Finally, under the supervision of the virtual normal vector and the reconstructed cloud point map, the surface normal vector and other 3D features are generated. Surface normal vector is an important parameter in 3D depth information, and the traditional surface normal vector calculation method is not robust. This paper proposes a surface normal vector calculation method with high robustness. If there are 2-dimensional pixel coordinates, VN will be mapped to the 3D space point coordinates with the following mapping formula:

$$
z_i=d_i,\ x_i=\cfrac{d_i(u_i-u_o)}{f_x},\ y_i=\cfrac{d_i(v_i-v_o)}{f_y}
$$

Where is the depth, which is the focal length of the x and y axes respectively. And is the center of the two-dimensional image lens. The acquisition is taken by a depth camera as a parameter of the data set.

The method of VNL sampling is to randomly select 3 * points from the "estimated depth map", a total of N groups, and the 3 points of each group are selected to meet the nonlinear and large-scale constraints. They are:

$$
\{ \alpha \geq \angle(\overrightarrow{P_AP_B},\overrightarrow{P_AP_C} \geq\beta, \alpha \geq \angle(\overrightarrow{P_BP_C},\overrightarrow{P_BP_A} \geq\beta, )\}
$$
$$
\{||\overrightarrow{P_kP_m}||>\theta,k,m \in group \}
$$

Among them, they are all hyperparameters. The above two formulas specify the angle relationship and distance requirements between the vectors of the same group. Generally speaking, the greater the distance, the more complete the global information, the closer the angle is orthogonal, and the same group will be more uncorrelated. In this way, based on simple mathematical analytic geometry knowledge, we can get a plane normal vector composed of 3 points in each group:

$$
\bold{n_i}=\cfrac{\overrightarrow{P_{Ai}P_{Ci}}\times \overrightarrow{P_{Ai}P_{Bi}}}{||\overrightarrow{P_{Ai}P_{Ci}}\times \overrightarrow{P_{Ai}P_{Bi}}||}
$$

The normal vector can be considered as a flat ID card. In this way we construct N normal vectors in the "estimated depth map. After calculation and derivation, it can be found that the robustness of the VN method is very high. It is assumed that the midpoint is biased due to noise, and the change of the normal vector is very small due to a wide range of conditions. VNL is the loss function for the data set that generated VN and trained above, and the expression is as follows:

$$
\mathcal{L}_{VN}=\frac{1}{N}(\Sigma_{i=0}^{N}||\bold{n_i}^{pred}-\bold{n_i}^{gt}||)
$$

In point-wise training, in order to make the system more integrated, we combine the point-wise loss function with VNL as the final VNL loss function.

$$
\mathcal{L}=\mathcal{L}_{WCE}+\lambda\mathcal{L}_{VN}
$$

is a hyperparameter, it has a trade-off mechanism, set to 5 in the experiment.
These are the basic principles of VNL. The core point of view is that when generating the normal vector, it is more "complete". It not only looks at the pixels and their neighboring points, but also adds the long-range constraint to improve the robustness of the algorithm. The introduction of nonlinear elements also guarantees functionality and efficiency. Our depth estimation results during experimental testing are shown in Figure 4.

\begin{figure}[htbp]
\centering
\subfigure[VNL]{
\begin{minipage}[t]{0.5\linewidth}
\centering
\includegraphics[width=1.5in]{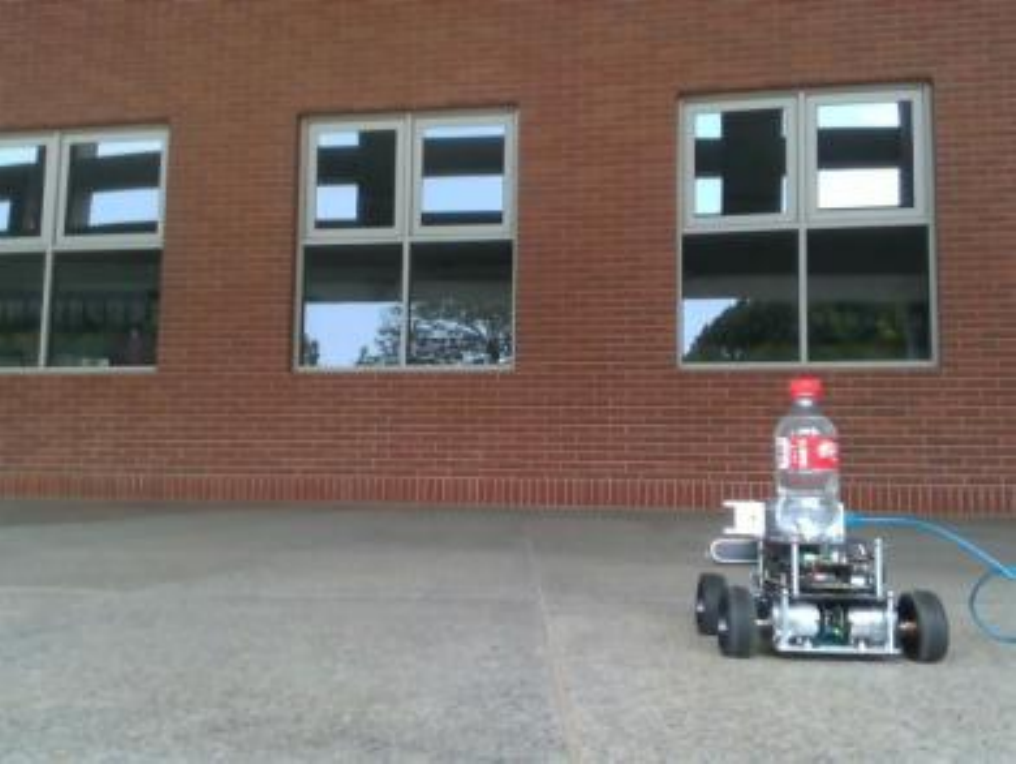}
\end{minipage}%
}%
\subfigure[Result]{
\begin{minipage}[t]{0.5\linewidth}
\centering
\includegraphics[width=1.5in]{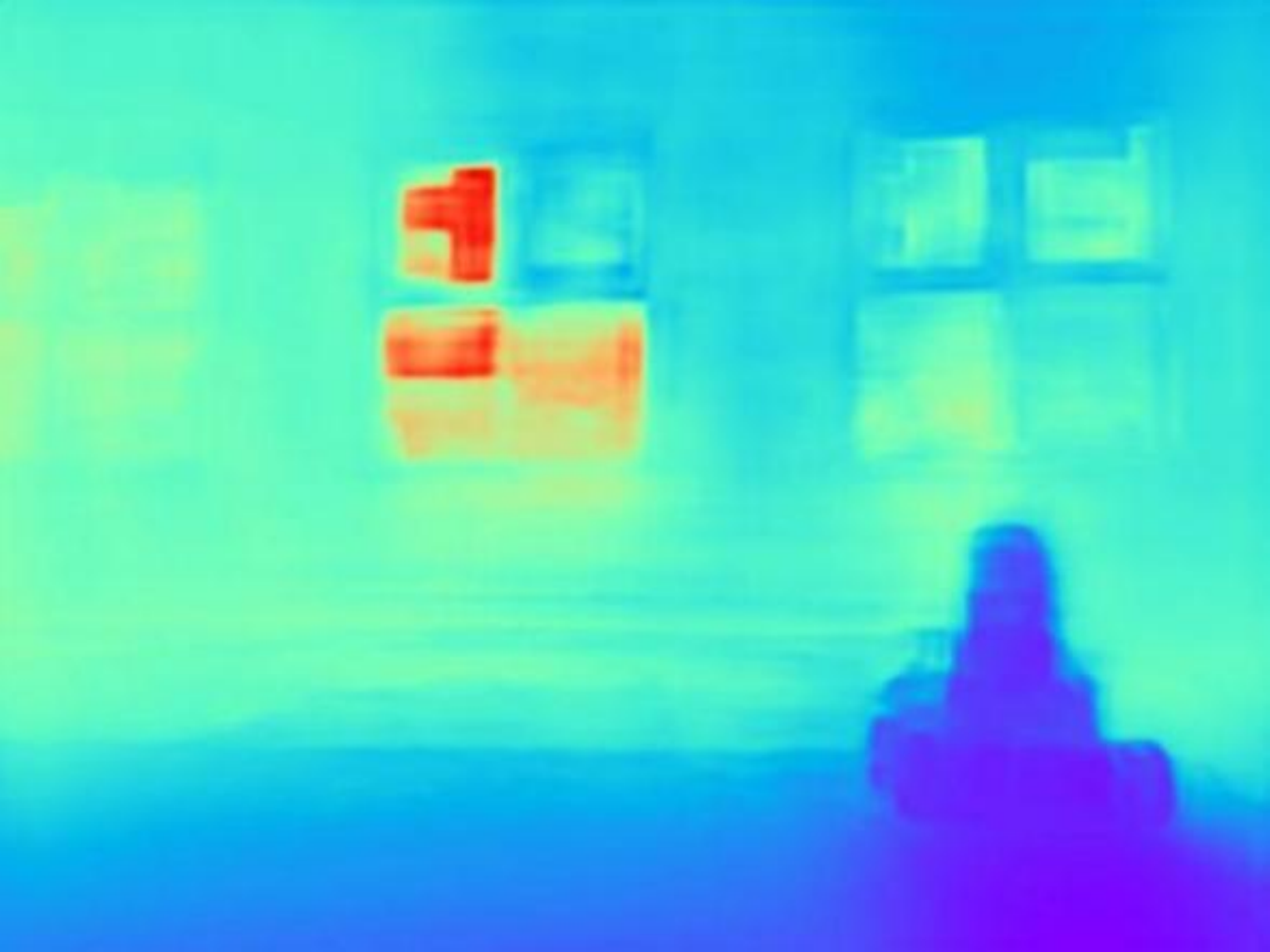}
\end{minipage}%
}%
\centering
\caption{VNL Experiment}
\end{figure}

\subsection{Fast estimation depth for PnP problem}
The method of using deep neural network has the disadvantage of poor timeliness. In our mission scenario, it is necessary to carry out depth detection on the target (i.e., the vehicle in front) in real time as much as possible to obtain the current position of the vehicle in front and quickly adjust its speed, orientation and other states accordingly. Therefore, in the time space when the neural network generates two frames of depth images, we use other faster methods to interpolate depth images.

In the task of following the car, our focus is on the relative position between the two cars, and we do not need to obtain the specific position information in the world coordinate system; therefore, we only need to know the coordinates of the vehicle ahead in the rear vehicle coordinate system. Based on the results of depth network reasoning, we can obtain continuous RGB images and intermittent depth maps. In visual SLAM, you can find the feature points in two RGB images for registration, and use the polar geometry method to estimate the camera pose change~\cite{daly1974polar}. Considering that in the current scene where only the relative position is concerned, this method can solve the change of the relative position of the two cars, but the problem is that monocular vision has scale uncertainty (Scale Ambiguity), and we have no way to get the exact value of the translation vector, but only the normalization result, which is inconsistent with our expected goal. Considering that we can obtain the depth data of some image frames through network output, we finally use PnP(Perspective-n-Point)~\cite{benito2008modularly} method to solve the motion of three-dimensional to two-dimensional points. This method describes how to estimate the location of the camera when $n$ three-dimensional space points and their projection positions are known.

Firstly, feature points are selected in three-dimensional space. Considering the high requirements of the algorithm for time and space efficiency, we look for feature points in two-dimensional images, and then calculate their position coordinates in three-dimensional space according to the depth map, thus obtaining three-dimensional space points. We have tried the classical SIFT~\cite{ng2003sift}, SURF~\cite{bay2006surf} and other feature point extraction algorithms~\cite{rublee2011orb}~`\cite{bay2008speeded} although this kind of method can find very good matching feature points, but its search and registration process time efficiency is not high. Through experiments, we finally chose a faster and more efficient Shi-Tomasi~\cite{bansal2021efficient} corner detection algorithm to carry out; As for the feature point matching part, considering that we are registering the images intercepted from the video stream, we chose the Lucas-Kanade~\cite{baker2004lucas} optical flow algorithm to complete the corner point matching between continuous frames. The results are shown in Figure 5.

\begin{figure}[htbp]
\centering
\subfigure[before the frame]{
\begin{minipage}[t]{0.33\linewidth}
\centering
\includegraphics[width=1.in]{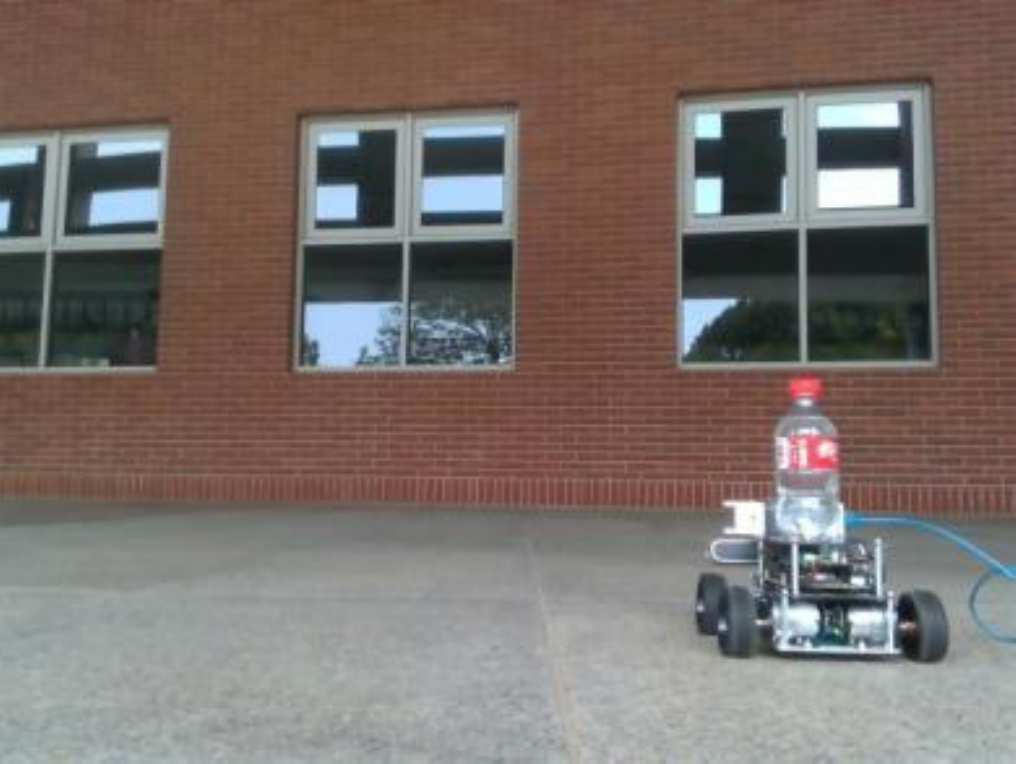}
\end{minipage}%
}%
\subfigure[Corner matching result]{
\begin{minipage}[t]{0.33\linewidth}
\centering
\includegraphics[width=1.in]{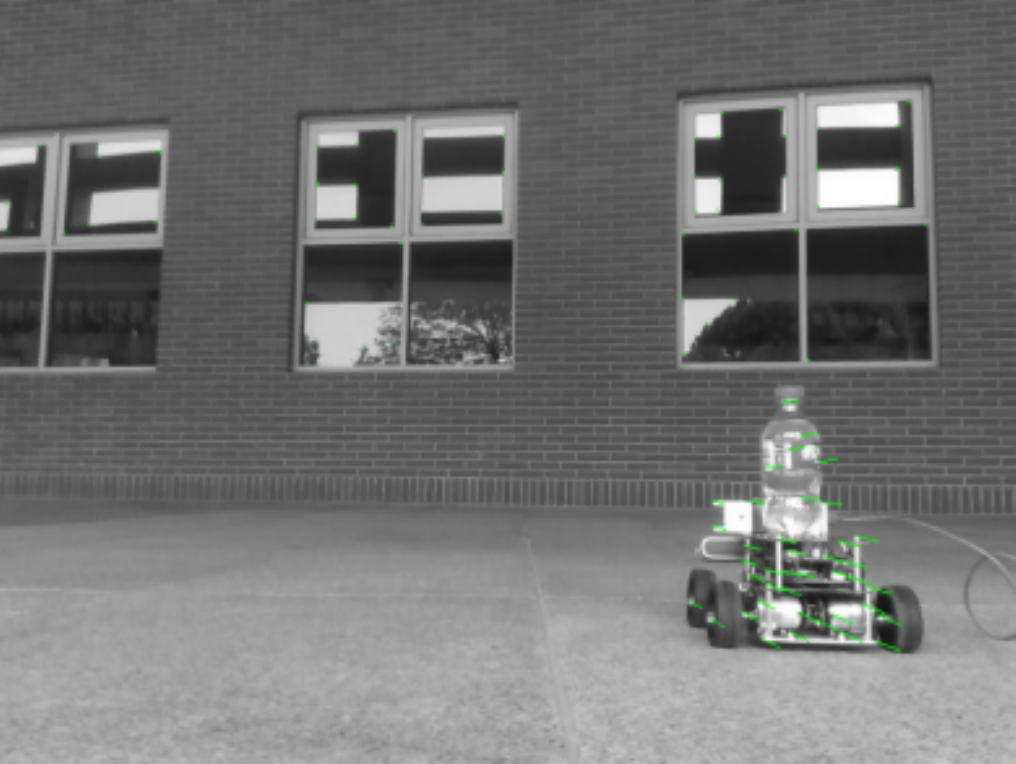}
\end{minipage}%
}%
\subfigure[after the frame]{
\begin{minipage}[t]{0.33\linewidth}
\centering
\includegraphics[width=1.in]{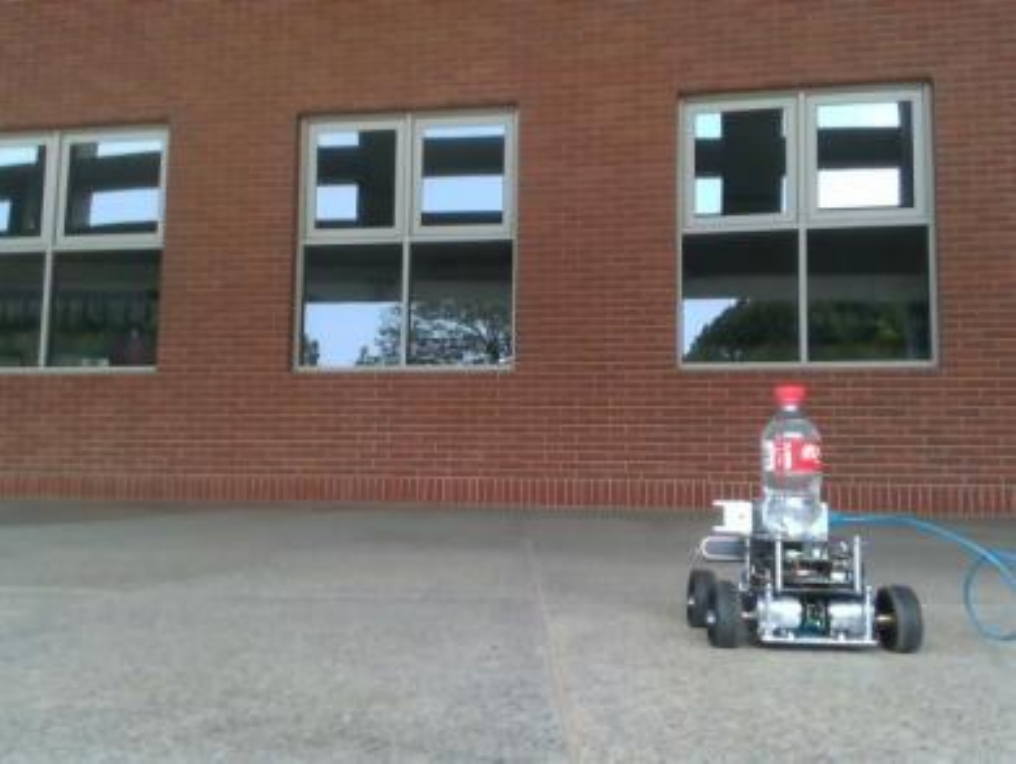}
\end{minipage}%
}%
\centering
\caption{Comparision}
\end{figure}

After selecting feature points, the feature points are classified. It is divided into two categories: foreground feature points and background feature points. Since we have completed the target detection task, we can think that the feature points in the target detection box are the upper points of the trolley, while the rest of the feature points are distributed in the background. Based on these two sets of point information, we can use the PnP method to calculate the relative pose change between the two cars and the pose change of the controlled car relative to the world coordinate system. Based on this, we can obtain the change of the distance between the two frames of images and roughly estimate the current speed of the controlled car.

When solving the PnP problem, we consider a certain space point, the expression in the homogeneous coordinate system is, at the same time in the image space, this three-dimensional point is projected. Calculate the change of camera position between the two frames of images, and remember that the rotation matrix and translation vector that characterizes this change and is unknown is, respectively,. Define the rotation matrix with the shape, expand it and write it:

$$
(R\ t)=
\left(
\begin{matrix}
t_1 & t_2 & t_3 & t_4 \\
t_5 & t_6 & t_7 & t_8 \\
t_9 & t_{10} & t_{11} & t_{12} \\
\end{matrix}
\right)
$$

According to the pinhole camera model, you can get:

$$
s\left(
\begin{matrix}
u_i\\
v_i  \\
1  \\
\end{matrix}
\right)=
\left(
\begin{matrix}
t_1 & t_2 & t_3 & t_4 \\
t_5 & t_6 & t_7 & t_8 \\
t_9 & t_{10} & t_{11} & t_{12} \\
\end{matrix}
\right)
\left(
\begin{matrix}
X\\
Y  \\
Z  \\
1
\end{matrix}
\right)
$$

Among them is a parameter, which can be eliminated by the last line of the equation system; it represents the position of the first feature point in the image. The first two equations can get two constraints. To simplify the expression, we define:
,,
You can get:
$$
\left\{
\begin{aligned}
T_1^TP-T_3^TPu_i&=0 \\
T_2^TP-T_3^TPv_i&=0 \\
\end{aligned}
\right.
$$

By sorting and superimposing the equations of n spatial feature points, we can list a set of linear equations:

$$
\left(
\begin{matrix}
P_1^T & 0 & -u_1P_1^T \\
0 & P_1^T & -v_1P_1^T \\
\vdots & \vdots & \vdots \\
P_N^T & 0 & -u_NP_N^T \\
0 & P_N^T & -v_NP_N^T \\
\end{matrix}
\right)
$$

The parameters to be solved are 12 dimensions in total, and each feature point can give 2 equations, so it should be noted that we need at least 6 pairs of matching points to achieve the solution of the pose. After experimental verification, it is found that it is easier to find 6 pairs of matching points when the car in front is not too far away. When this indicator is not met, P3P~\cite{cranor2003p3p} and other methods can be tried.

\subsection{Information Fusion}
First extract the position of the trolley from the depth map. Since the front vehicle mark obtained in the target detection is a rectangular frame, which will be mixed with background components, the average depth in this area cannot well characterize the distance information between the two vehicles. Observe the depth distribution in a typical rectangular frame, as shown in Figure 2-8. It can be seen that there will be a peak in the depth density, and the depth corresponding to this peak can be considered as the depth of the vehicle ahead in the image. According to this, the steps we chose to obtain the depth are: 1)divide the depth in the rectangular box of the marked vehicle into 50 parts, obtain its probability density distribution, find the peak value among them 2) expand the depth range of the peak value to the left and right, and calculate its average value to obtain the final estimated depth.

The estimated depth can be roughly obtained by performing the first step, but considering the possible influence of noise, we extend the depth range and calculate the average value to obtain more accurate depth information. At the same time, we can use a statistical-based method to distinguish the foreground and background using the sigma principle. The results are shown in Figure 6(a). The red is the divided foreground area, and the blue is the background area.

\begin{figure}[htbp]
\centering
\subfigure[Depth Density]{
\begin{minipage}[t]{0.5\linewidth}
\centering
\includegraphics[width=1.5in]{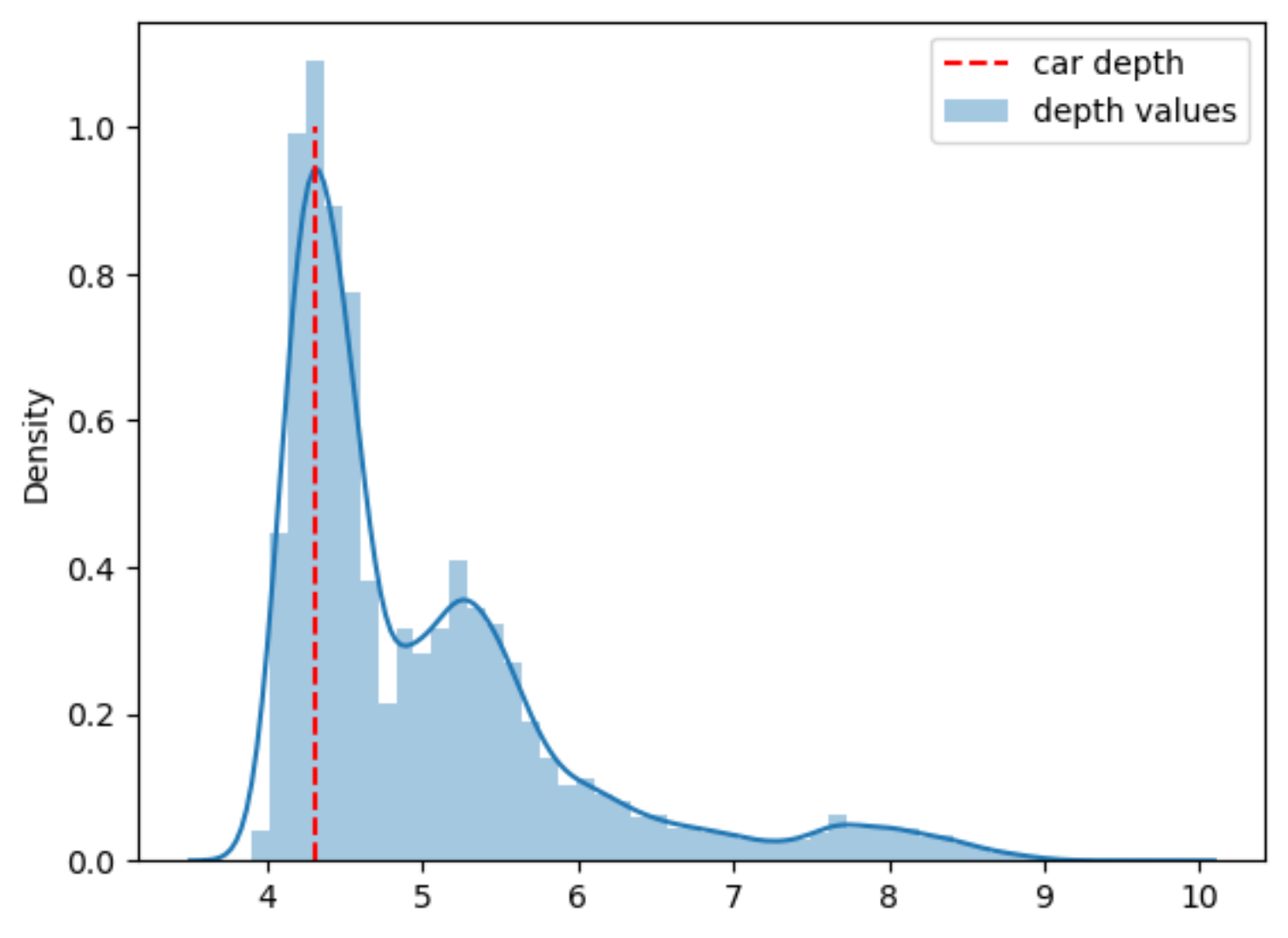}
\end{minipage}%
}%
\subfigure[Results of foreground background division]{
\begin{minipage}[t]{0.5\linewidth}
\centering
\includegraphics[width=1.5in]{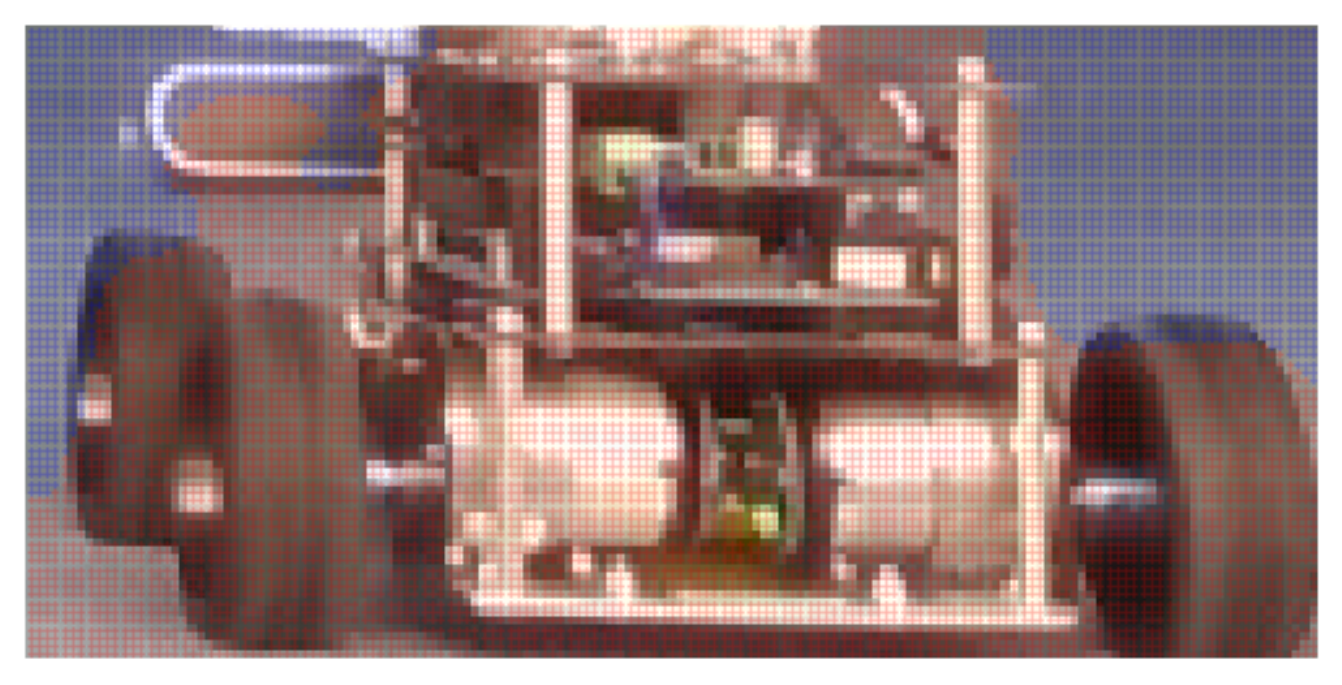}
\end{minipage}%
}%
\centering
\caption{Experiment}
\end{figure}

After obtaining more accurate depth information, we believe that the center of the rectangular frame is the center coordinate of the vehicle in front in the image. According to the pinhole camera model, the pixel points on the image have the following relationship with the positions of the three-dimensional coordinate points in the world:

$$
Z\left(
\begin{matrix}
u\\
v\\
1
\end{matrix}
\right)
=
K
\left(
\begin{matrix}
u\\
v\\
1
\end{matrix}
\right),
K=
\left(
\begin{matrix}
f_x & 0 & c_x\\
0 & f_y & c_y\\
0 & 0 & 1
\end{matrix}
\right)
$$
where is the homogeneous coordinate expression of the pixel in the image coordinate system, is the camera internal parameter matrix, and is the corresponding three-dimensional coordinate point spatial position, that is, the point depth. Knowing the camera's internal parameter matrix and pixel coordinates, plus the estimated depth information, you can get the space position of the front car. The calculation method is:
$$
\left\{
\begin{aligned}
X&=\cfrac{(u-c_x)Z}{f_x} \\
Y&=\cfrac{(v-c_y)Z}{f_y} \\
\end{aligned}
\right.
$$
At this point, we have completed the depth estimation of the position of the vehicle ahead, and then we need to fuse the depth information.

The methods used in depth estimation are deep neural network and PnP method. Compared with traditional sensors, the lag between signal sampling time and transmission reaching time cannot be ignored. Therefore, the control signal and the two depth information signals are out of step, and we need to calibrate this. We used the method of the Kalman filter~\cite{welch1995introduction}~\cite{meinhold1983understanding} to obtain an optimal estimate of the depth. State variables are selected, where the variables characterize the three-dimensional spatial position and velocity of the vehicle ahead, respectively. We store part of the historical position information in the buffer area, and use the cubic polynomial to fit the short-term trajectory of the front car obtained from the estimated depth of PnP, and correct the motion to a certain extent. Considering that the modeling process is a discrete system, we define the following state transfer equation:
$$
R= \cfrac{d^2_{depth}-d^2_{exp}}{2x}, \delta = tan^{-1}\cfrac{2xL}{d^2_{depth}-d^2_{exp}}
$$

where is the process noise covariance matrix. It can be seen that fitting the position information with a cubic polynomial means that the three derivatives of time are continuous, that is to say, the acceleration is continuously changing, and the trajectory of the vehicle ahead will be smoother.
The depth information we inferred for each frame is timestamped. Since the PnP depth estimation signal is generated more frequently, we use it as a benchmark to register the network depth estimation information on it. Whenever the PnP estimates a depth map, we store its time and label in the cache area. The specific correction method is to calculate the difference between the estimated coordinate result of the frame network and the estimated coordinate result of PnP to obtain an error term. At the same time, considering that the depth map deduced by the PnP method is based on the frame, this error term is propagated to the following position estimation to obtain the corrected result. This repair is being carried out in the location information buffer area. At this time, it is necessary to refit the trajectory polynomial, which in turn has an effect on the current and subsequent front vehicle position estimation.

\subsection{Control Method}
The control framework diagram of the system is shown in Figure 7. Among them, the target detection node detects the moving target through the information subscribed by the camera, outputs the size and position of the detection box and transmits it to the depth estimation node and the motion control node. The depth estimation node calculates the depth information of the target and transmits it to the motion control node. The motion control node receives the position information of the detection box and the target depth information, and calculates the corresponding speed and direction control amount. In the target detection part, the lateral position offset of the target can be obtained by detecting the deviation between the center of the frame and the axis in the camera field of view. The depth estimation part will feed back the depth information, which can be used to approximate the distance between the target and the trolley.

The desired control result is that the direction of the moving platform is always towards the moving target and maintains a fixed distance from it. Using the idea of pure tracking method [15], for a trolley with Ackerman steering structure, when the direction angle is time, the trolley will move in a trajectory with a radius. Among them, L is the length of the body. Assuming that A is our moving target, B is the car's preview point. According to the path of Figure 2-11, in the case of known lateral position offset and depth information, you can further obtain the direction angle:
\begin{figure}[htbp]
\centering
\subfigure[Process]{
\begin{minipage}[t]{0.5\linewidth}
\centering
\includegraphics[width=1.5in]{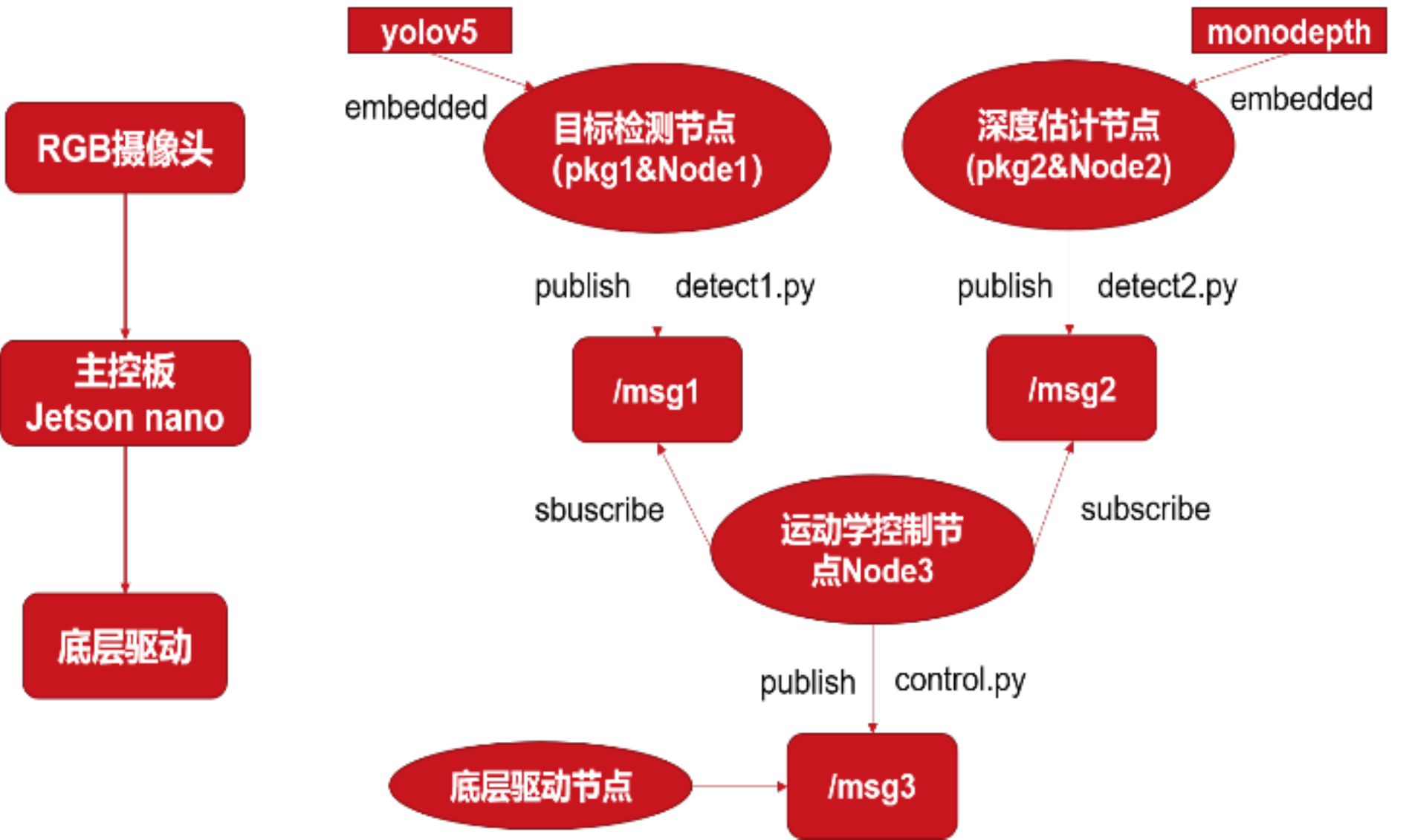}
\end{minipage}%
}%
\subfigure[Theory]{
\begin{minipage}[t]{0.5\linewidth}
\centering
\includegraphics[width=1.5in]{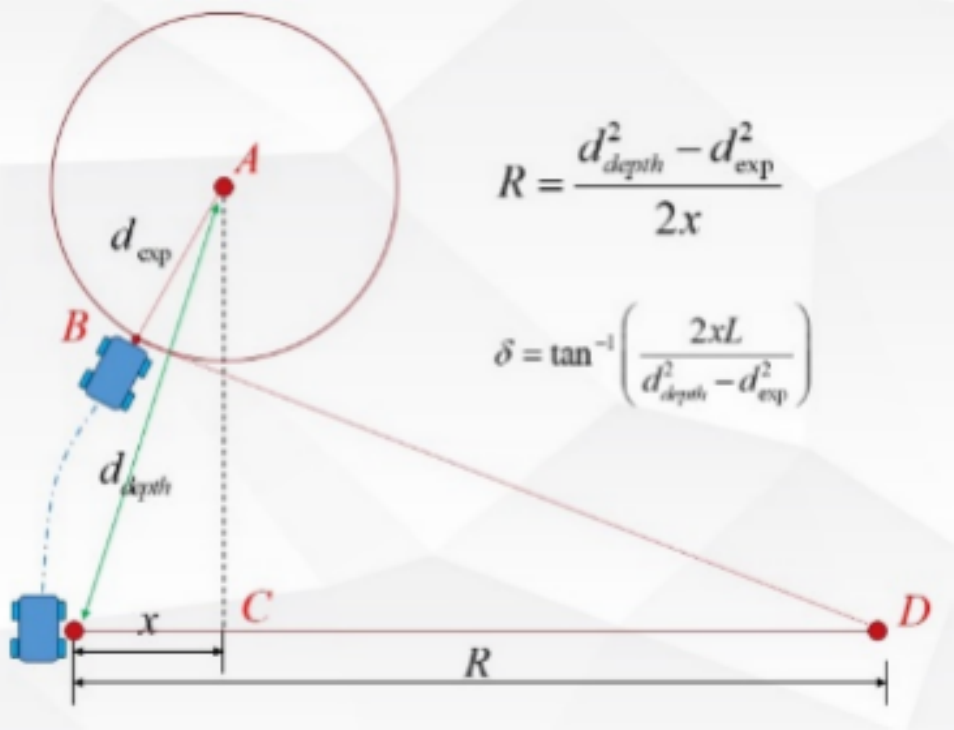}
\end{minipage}%
}%
\centering
\caption{Control}
\end{figure}

When approaching, the denominator of the direction angle approaches 0, and the angle approaches 90 °. Therefore, the actual output angle needs to be limited. PID control is used for speed, and the difference between the current depth of the trolley and the expected depth is used as the error. When the car moves to within a circle with a radius of, the issuing speed is 0 to make the car stationary.

\section{Results}
Based on the above methods, we successfully deployed the YOLO-v5 and VNL Depth Estimation on the mobile platform controlled by the Jetson-nano, and can realize the tracking control of specific low-speed moving targets according to the image information. The target detection part can successfully identify the moving target, and the depth estimation part can return a stable depth map. The research results are helpful to carry out more in-depth research on embedded platform.

In the target detection part, our reasoning speed for each frame reaches 0.01s, and experiments show that the prediction results have high accuracy.

In the depth estimation section, Tables 2 show the experimental results based on depth learning and PnP depth estimation methods, respectively. Here, the reasoning results of the original VNL method are used as the standard and compared with our proposed deep information fusion method. The experimental results show that compared with neural network reasoning, the reasoning speed of deep information fusion is increased by three times and the accuracy is improved by 13\%. The calculation formula of some of the indicators is as follows, where is the total number of pixels, is the depth true value image, is the depth estimation image.

$$
\begin{aligned}
E_{abs} &= \cfrac{1}{P} \sum^{P}_{i=1}\mid\cfrac{gt(i)-pred(i)}{gt(i)}\mid \\
E_{sq} &= \cfrac{1}{P} \sum^{P}_{i=1}\mid\cfrac{gt(i)-pred(i)}{gt(i)}\mid^2 \\
E_{rms} &= \sqrt{\cfrac{1}{p}\sum^P_{i=1}[gt(i)-pred(i)]^2}\\
E_{logrms} &= \sqrt{\cfrac{1}{p}\sum^P_{i=1}[loggt(i)-logpred(i)]^2}
\end{aligned}
$$

\begin{table}[H]
	\centering
	\caption{\textbf{VNL vs PnP results}}
	\footnotesize
	\setlength{\tabcolsep}{6pt}
	\begin{tabular}{ccccccc}
		\toprule [1.3pt]	
		& Time(s)& Abs& Sq& RMS& logRMS& Accuracy  \\
		\hline
		VNL&0.2721&0.1059&0.0278&0.4071&0.1418&0.8825 \\
		\hline
		PnP&0.0886&0.0359&0.025&0.5229&0.0478&0.9975 \\
		\hline
		\bottomrule[1.3pt]
	\end{tabular} \label{table:TableI}
\end{table}

After deep information fusion, Kalman filter test is performed. The results are shown in Figure 8\&9. The green curve is the interpolation result of PnP estimation method, the blue curve is the depth calculated by the neural network output, and the red curve is the result of Kalman filtering by the above method. It can be seen that after a period of adjustment, the filter gradually approaches convergence, and can give more accurate and stable depth prediction results, while meeting the needs of difference and jitter elimination.

\begin{figure}[H]
    \centering
    \includegraphics[width=8cm]{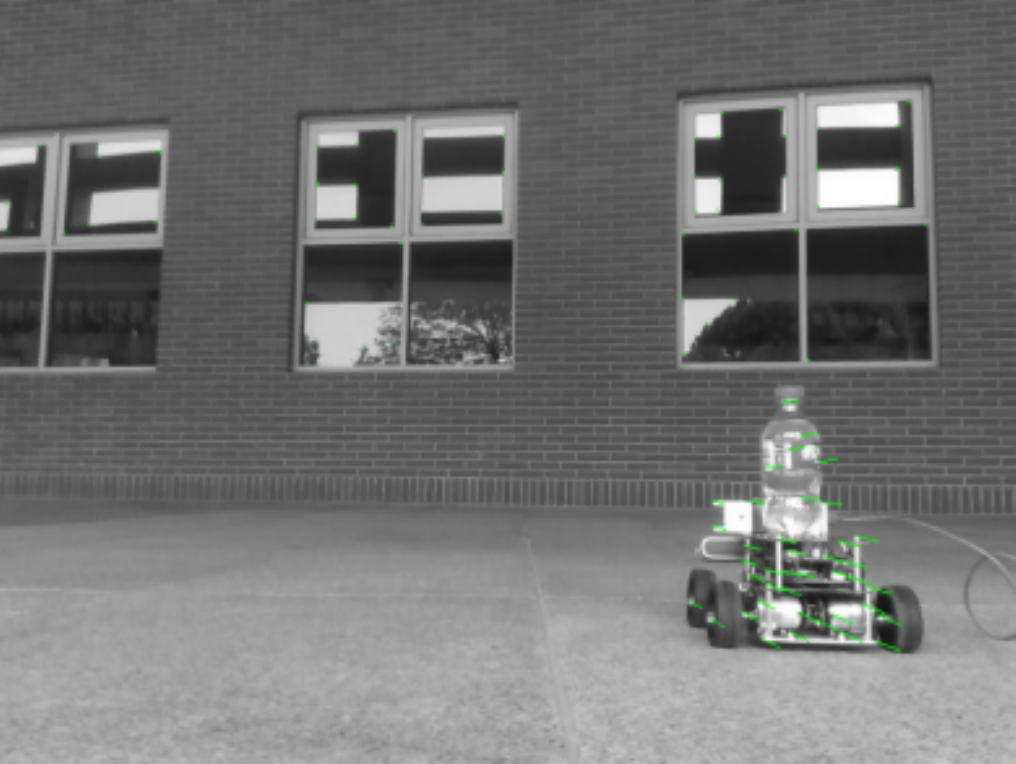}
    \caption{After Filter}
    \centering{\label{fig:my_label}}
\end{figure}

\begin{figure}[H]
    \centering
    \includegraphics[width=8cm]{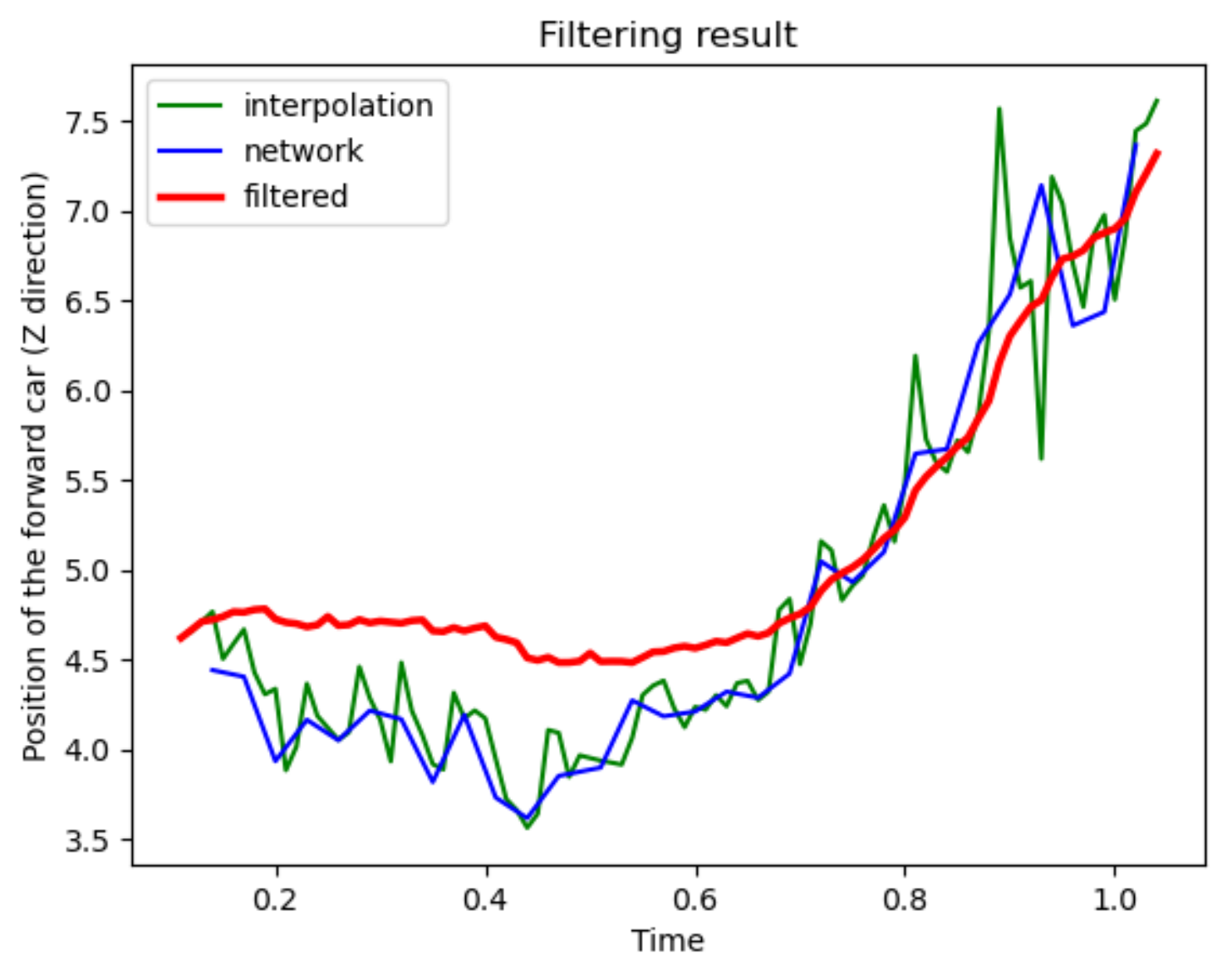}
    \caption{After Filter}
    \centering{\label{fig:my_label}}
\end{figure}

\section{CONCLUSIONS}
RealNet can achieve high rate of accuracy and speed when recognizing objects and depth information on a real-time device like IoT, etc. The experiment results are idea, which is 0.01s for object detection using optimized YOLO-v5 and 3 times faster than pure depth estimation neural network methods. We add some information fusion strategy in it and perform some classic methods like Kalman filter to help it better performance.

We're about to release some new versions of the system after this paper published.

\bibliographystyle{plain}
\bibliography{s.bib}

\end{document}